\documentclass[10pt]{article}
\usepackage[preprint]{tmlr}

\usepackage{amsmath,amsfonts,bm}

\def\eqref#1{equation~\ref{#1}}

\def\1{\bm{1}}

\DeclareMathAlphabet{\mathsfit}{\encodingdefault}{\sfdefault}{m}{sl}
\SetMathAlphabet{\mathsfit}{bold}{\encodingdefault}{\sfdefault}{bx}{n}

\newcommand{\softmax}{\mathrm{softmax}}

\usepackage{hyperref}
\usepackage{url}

\usepackage{microtype}
\usepackage{graphicx}
\usepackage{subcaption}
\usepackage{booktabs} 
\usepackage{xcolor}   
\usepackage{makecell}

\usepackage{amsmath}
\usepackage{amssymb}
\usepackage{mathtools}
\usepackage{amsthm}
\usepackage{enumitem}
\setlist{nosep}

\usepackage{multirow}  

\usepackage{wrapfig}

\definecolor{mygreen}{rgb}{0.15,0.55,0.25}
\definecolor{AlgoGreen}{RGB}{25,135,60}
\definecolor{algcmt}{rgb}{0.35,0.35,0.35}

\usepackage{algorithm}

\let\saveTmlrAND\AND
\let\AND\relax
\usepackage{algorithmic}
\makeatletter
\let\orig@algorithmic\algorithmic
\def\algorithmic{%
  \let\AND\relax
  \orig@algorithmic
}
\makeatother

\newcommand{\algCommentCol}{0.40\linewidth} 
\newcommand{\algCmtR}[1]{%
  \hspace*{\fill}\parbox[t]{\algCommentCol}{\raggedright\footnotesize\textcolor{mygreen}{//~#1}}%
}
\providecommand{\algCmt}[1]{\algCmtR{#1}}

\providecommand{\IMG}{\texttt{[IMG]}}
\providecommand{\eg}{e.g.,\ }

\newcommand{\para}[1]{\noindent\textbf{#1.}\,}
\newcommand{\TopK}{\operatorname{TopK}}
\newcommand{\GAD}{\textsc{GAD}}
\newcommand{\LGD}{\textsc{LGD}}
\newcommand{\avgup}[1]{\textcolor{mygreen}{\scriptsize$\uparrow$\,#1}}
\newcommand{\avgdown}[1]{\textcolor{red}{\scriptsize$\downarrow$\,#1}}

\usepackage[capitalize,noabbrev]{cleveref}

\theoremstyle{plain}

\theoremstyle{definition}

\theoremstyle{remark}

\newcommand{\upgain}[1]{%
  \smash{\rlap{\hspace{0.1em}\raisebox{0.8ex}{%
    \scriptsize\textcolor{mygreen}{$\uparrow$\,#1}%
  }}}%
}

\title{Geometry-Aware Distillation for Prompt Tuning Biomedical Vision-Language Models}

\let\AND\saveTmlrAND
\author{\name Tran Dinh Tien \email tien.tran@mbzuai.ac.ae \\
      \addr Department of Machine Learning \\
      Mohamed bin Zayed University of Artificial Intelligence
      \AND
      \name Zhiqiang Shen\textsuperscript{\dag} \email zhiqiang.shen@mbzuai.ac.ae \\
      \addr Department of Machine Learning \\
      Mohamed bin Zayed University of Artificial Intelligence}

\begin{document}

\maketitle
\renewcommand{\thefootnote}{\fnsymbol{footnote}}
\footnotetext[2]{Corresponding author.}
\renewcommand{\thefootnote}{\arabic{footnote}}

\begin{abstract}
Current prompt-based and adapter-based tuning of vision-language models (VLMs) is attractive for medical imaging, where clinical data sensitivity favors frozen backbones and annotations are limited. However, these methods typically optimize only the ground-truth class, treating all other classes as equally incorrect, ignoring clinically meaningful class relations and yielding unstable decision boundaries in limited-supervision settings. We propose Omni-Geometry Knowledge Distillation (OGKD), a new framework that injects class-relation structure into the teacher to produce directional targets that preserve the ground truth while respecting inter-class geometry. Using these targets, we develop two distillation losses: Global Geometry-Aware Distillation (GAD) operates on the global image token, and Label-Guided Geometry Distillation (LGD) applies the same geometry to attentive patch tokens to improve fine-grained alignment. Across comprehensive experiments and analyses on 11 widely-used medical datasets for base-to-novel and few-shot evaluations, our OGKD achieves substantially better performance, consistently improving accuracy by an average absolute gain of {\bf 1.7\%–2.8\%} over all prior state-of-the-art VLM adaptation counterparts. It also robustly generalizes to unseen classes and yields more reliable predictions than other approaches. Our code is available at \url{https://github.com/tientrandinh/OGKD}.
\end{abstract}

\section{Introduction}
Annotated clinical imaging data are often scarce, particularly for rare conditions and novel acquisition protocols~\citep{litjens2017survey}. This motivates methods that generalize from minimal supervision. Few-shot learning addresses this setting by enabling generalization from limited labeled samples~\citep{vinyals2016matching,snell2017prototypical,finn2017model}. When combined with vision--language pretraining, models can condition on textual descriptors, which is a powerful route to improve data efficiency~\citep{radford2021learning,jia2021scaling,zhai2022lit}. In medical imaging, domain-specific vision--language models (VLMs) show that this route can unlock broad transfer with limited labels~\citep{wang2022medclip,zhang2023biomedclip,bannur2023learning}. These developments make few-shot adaptation of biomedical VLMs a promising direction for building data-efficient diagnostic tools.
\begin{figure}[t]
  \centering
  \includegraphics[width=0.8\linewidth]{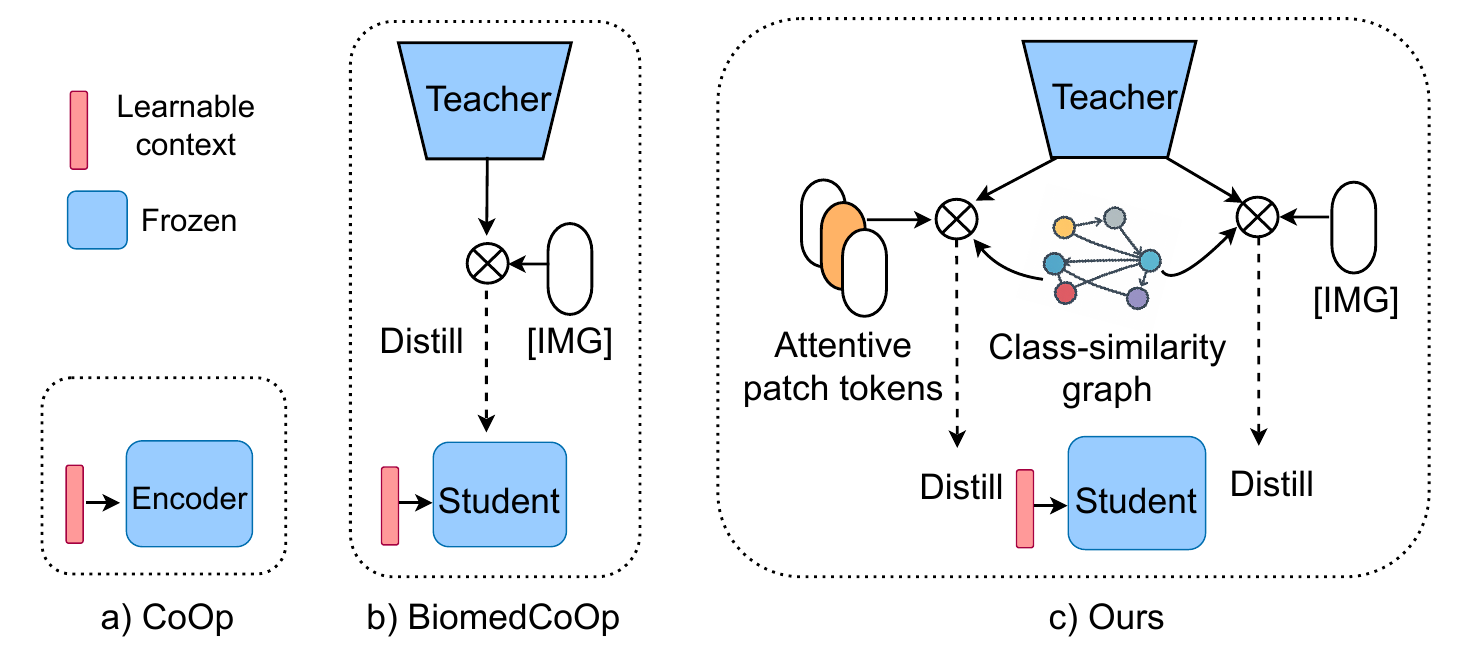}
  \caption{Comparison of prompt-learning pipelines: (a) CoOp~\citep{zhou2022learning}, (b) BiomedCoOp~\citep{koleilat2025biomedcoop}, and (c) Ours. Prior methods typically treat classes independently and rely mainly on global image representations; our approach leverages class semantic structure by shaping the teacher distribution with a class graph \(W\) and distills geometry-aware supervision over both global \IMG{} token and attentive patch tokens. Only prompts are learned; the encoders remain frozen.}
  \label{fig:overview-compare}
\end{figure}

\begin{figure}[t]
  \centering
  \includegraphics[width=1.0\linewidth]{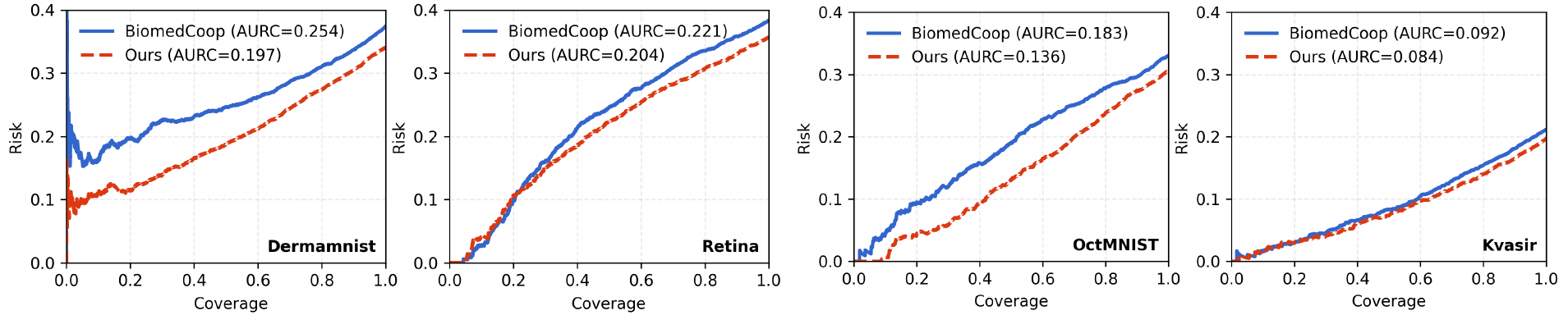}
  \caption{\textbf{Risk--coverage across four biomedical datasets.} Our method yields consistently lower curves (smaller AURC) than BiomedCoOp~\citep{koleilat2025biomedcoop}, indicating fewer high-confidence errors at the same coverage and improved selective reliability.}
  \label{fig:aurc_intro}
\end{figure}

Adapting pretrained VLMs typically follows two directions. (1) \emph{Prompt learning} optimizes lightweight textual prompts while freezing the backbone; representative methods include CoOp~\citep{zhou2022learning}, CoCoOp~\citep{zhou2022conditional}. (2) \emph{Adapter-based} tuning adds small trainable modules on top of a frozen backbone, including Tip-Adapter~\citep{zhang2021tip}, CLIP-Adapter~\citep{gao2024clip}; linear probing~\citep{radford2021learning,huang2024lp++} is also widely used as a strong baseline. Recently, knowledge distillation (KD) methods~\citep{hinton2015distilling} have also been introduced for CLIP-style models~\citep{yang2024clip,li2024promptkd}, but they typically require heavy student fine-tuning or two-stage training with additional layers.

Biomedical imaging poses additional challenges due to complex visual features, diverse modalities, and heterogeneous acquisition devices. BiomedCoOp~\citep{koleilat2025biomedcoop} uses LLM-generated templates as teacher prefix prompts and trains student prompts by transferring knowledge from the teacher, achieving strong performance on biomedical datasets. However, existing prompt-, adapter-, and KD-based methods typically (i) treat non-target classes uniformly, ignoring semantic relations (e.g., \emph{benign} is closer to \emph{malignant} than \emph{malignant} is to \emph{normal}); and (ii) distill only the global \IMG{} token, missing fine-grained information at the patch-token level.

In safety-critical domains such as medical diagnostics, models must also know when to abstain~\citep{geifman2017selective}. The risk-coverage curve in Fig.~\ref{fig:aurc_intro} measures selective reliability: as coverage decreases (keeping only the most confident predictions), risk (error among retained samples) should drop quickly, yielding a small area under the risk--coverage curve (AURC). Empirically, we find that strong methods such as BiomedCoOp~\citep{koleilat2025biomedcoop}, which emphasize the ground-truth logit while treating all non-targets as equally incorrect, can produce overconfident mistakes and increase AURC. Our method shifts probability mass toward semantically related classes and reduces spurious confidence in distant ones, lowering AURC and complementing accuracy gains. Recent work~\citep{fan2024seeing} explores geometry in biomedical settings for novel concept discovery rather than supervised few-shot VLM adaptation. More broadly, recent VLM adaptation methods have begun to model inter-class structure through graph-based modules or similarity-preserving prompt regularization~\citep{li2023graphadapter,jung2025learning}. Our setting is different: rather than learning additional graph/adaptation modules or regularizing prompt embeddings directly, we construct a fixed text-derived class graph and use it to shape teacher targets for few-shot biomedical prompt tuning.

Medical VLMs have also explored global--local alignment between image regions and report tokens (e.g., GLoRIA~\citep{huang2021gloria}, LoVT~\citep{muller2022joint}), supporting the need for local supervision beyond global embeddings. In this work, we focus on few-shot prompt tuning with geometry-aware teacher shaping at both the global \IMG{} token and label-guided patch-token levels (Fig.~\ref{fig:overview-compare}). 

\noindent\textbf{Motivation.}
Expert clinicians rarely weigh a target diagnosis equally against all alternatives. A radiologist might form an initial impression of an anomaly and then focus on a narrow differential diagnosis rather than exhaustively evaluating every option~\citep{krupinski2010current}. Existing models, however, often treat non-targets uniformly despite strong semantic relations among medical classes. Motivated by this gap, we introduce an analogous prior for VLMs: when supervision is limited, errors should fall on semantically related classes rather than unrelated ones.

Our contributions are summarized as follows:
\begin{itemize}
\item We encode class-relation structure into the VLM distillation framework by constructing a \emph{text-driven} class graph \(W\) from frozen biomedical text prototypes, without using visual samples. This mitigates overfitting in few-shot regimes while injecting clinically meaningful inter-class geometry.
\item We introduce an Omni-Geometry Knowledge Distillation method (OGKD) with two losses: (i) \emph{Global Geometry-Aware Distillation (GAD)} shapes the teacher distribution at the global \IMG{} token and distills the full class distribution; and (ii) \emph{Label-Guided Geometry Distillation (LGD)} applies the same geometry at label-guided patch tokens and distills the label channel to emphasize fine-grained alignment. The approach is parameter-efficient: only student prompts are learned.
\item Across 11 biomedical datasets, OGKD delivers consistent gains over state-of-the-art VLM adaptation methods in base$\rightarrow$novel generalization, few-shot classification, and selective reliability (risk--coverage), indicating both higher accuracy and clinically safer decisions.
\end{itemize}

\section{Related Work}
\label{sec:related}
\para{Vision--language Models (VLMs)}
Large-scale VLMs such as CLIP~\citep{radford2021learning} and ALIGN~\citep{jia2021scaling} learn open-vocabulary recognition by contrasting images and texts at scale. Although highly effective on natural images, multiple works have documented a domain gap when transferring these models to medical images, motivating domain-specific pretraining~\citep{boecking2022making,eslami2021does,zhang2023biomedclip}. BioViL~\citep{boecking2022making} aligns radiology reports with images. PubMedCLIP~\citep{eslami2021does} and BiomedCLIP~\citep{zhang2023biomedclip} scale up with curated biomedical pairs. These works improve zero/few-shot transfer to medical images, but they still struggle with fine-grained categories due to limited localized supervision and medical data imbalance~\citep{boecking2022making,zhang2023biomedclip}.

Recent work has also developed specific medical vision--language foundation models, such as FLAIR~\citep{silva2025foundation} for ophthalmology, as well as PLIP~\citep{huang2023visual}, QuiltNet~\citep{oluchi2023quilt}, and CONCH~\citep{lu2024visual} for pathology. These efforts highlight the impact of foundation models for healthcare, but downstream use still requires reliable adaptation to new label spaces and modality shifts, especially across settings like our benchmark spanning nine modalities and ten anatomical sites. We therefore focus on a parameter-efficient distillation framework that transfers a frozen biomedical VLM across diverse domains without training a new foundation model per specialty.

\para{Prompt learning and adapters}
Prompt learning adapts VLMs by optimizing a small set of textual tokens instead of all model weights. CoOp~\citep{zhou2022learning} learns class-specific prompts and achieves strong accuracy, but subsequent studies (e.g., CoCoOp~\citep{zhou2022conditional} and KgCoOp~\citep{yao2023visual}) show that naive prompt tuning can \emph{overfit} to base classes and generalize poorly under distribution shift. MaPLe~\citep{khattak2023maple} and PromptSRC~\citep{khattak2023self} argue that using both visual and textual prompts can mitigate this instability. KgCoOp~\citep{yao2023visual} constrains prompts with a knowledge-guided objective to preserve CLIP priors. ProGrad~\citep{zhu2023prompt} regularizes gradients to avoid negative transfer. Another direction is to use \emph{adapter} modules: CLIP-Adapter~\citep{gao2024clip} inserts lightweight residual adapters, while Tip-Adapter~\citep{zhang2021tip} caches features to enable training-free or fast tuning. LP++~\citep{huang2024lp++} improves linear probing with strong regularization. These methods reduce trainable parameters but operate on \emph{global} image tokens and thus miss fine-grained signals.

\para{Inter-class relations in VLM adaptation}
Traditional prompt tuning typically treats categories as independent targets, which can overlook the semantic relationships originally captured by pretrained VLMs. Recent work has been exploring inter-class structure to regularize adaptation. For example, GraphAdapter~\citep{li2023graphadapter} is an adapter-style method that models class relations with dual knowledge graphs built from textual and visual structure, while SAR~\citep{jung2025learning} regularizes learnable prompts so that pairwise similarities among prompt-induced text embeddings remain aligned with those from hand-crafted prompts, using LLM-generated related novel classes as auxiliary anchors.

Our OGKD framework shares the high-level goal of preserving class semantics, but differs in mechanism and supervision granularity. Unlike GraphAdapter~\citep{li2023graphadapter}, which introduces additional trainable adapter and graph-learning components, and unlike SAR~\citep{jung2025learning}, which directly regularizes prompt-induced text embeddings using auxiliary LLM-generated classes, OGKD (ours) constructs a fixed class graph \(\mathbf{W}\) once from frozen biomedical text prototypes over the predefined label set. This design is intentional for few-shot biomedical settings, where estimating class relations from limited visual samples can be unstable and may overfit dataset-specific artifacts or annotation noise. In addition, while these methods operate exclusively at the global feature level, OGKD utilizes an \emph{omni-geometry} approach that projects this class structure down to \emph{label-guided attentive patch tokens} (\cref{sec:omni-attn}). This local distillation is essential for biomedical imaging, where diagnostic evidence is highly localized.

\para{Prompt learning in biomedical imaging}
BiomedCoOp~\citep{koleilat2025biomedcoop} applies knowledge distillation~\citep{hinton2015distilling} with a medical encoder backbone~\citep{zhang2023biomedclip} and shows competitive performance. In line with most prompt-tuning methods, it (i) treats classes independently and (ii) distills only at the \emph{global} token level. As widely observed in medical recognition, relying only on global signals can miss discriminative local pathology cues. In contrast, we address these limitations by (i) exploiting semantic structure among classes, and (ii) introducing \emph{omni-geometry distillation} that preserves geometry-aware structure at both the global token and label-guided local tokens. 
\begin{figure}[t]
  \centering
  \includegraphics[width=\textwidth]{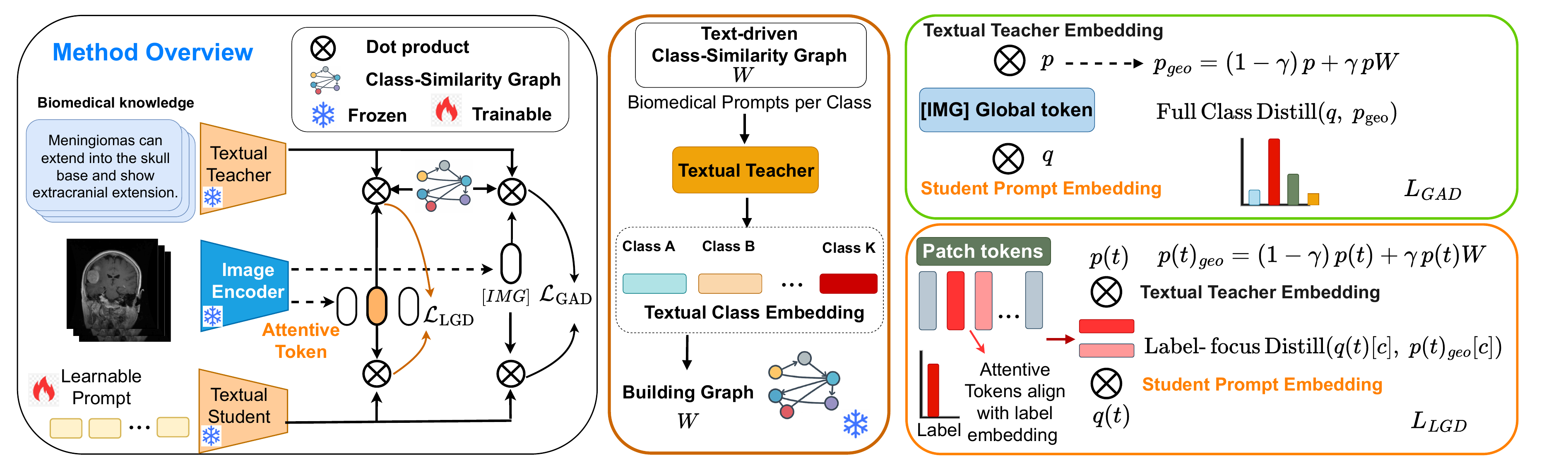}
  \caption{From biomedical text prototypes, we build a class graph \(\mathbf{W}\) that captures semantic relations among classes. A geometry strength \(\gamma\) smooths the teacher distribution and supervises two distillation losses: (1) \textbf{GAD} operates at the global \IMG{} token; (2) \textbf{LGD} operates at patch-token level, where \(c\) denotes the ground-truth class. Only the student prompts are updated; the encoders and \(\mathbf{W}\) remain frozen.}
  \label{fig:detail-method-wide}
\end{figure}

\section{Method}
\para{Preliminaries}
We are given a few-shot training set $\mathcal{D}=\{(\mathbf{x}_i,y_i)\}_{i=1}^{N}$ with classes $\mathcal{C}=\{1,\dots,C\}$. We build on a frozen VLM~\citep{zhang2023biomedclip} with an image encoder $g^{\mathrm{img}}$ and a text encoder $g^{\mathrm{text}}$. Let $d$ denote the shared embedding dimension. Following prompt learning, we optimize only a small learnable context $\mathbf{C}\in\mathbb{R}^{n_{\text{ctx}}\times d}$ inserted into the text prompt; all encoder parameters are frozen. Given an image $\mathbf{x}$, the global image token $\mathbf{v}(\mathbf{x})\in\mathbb{R}^{d}$ is the projected \IMG{} token produced by $g^{\mathrm{img}}$ and is L2-normalized. For a class $c\in\mathcal{C}$, the textual prototype $\mathbf{t}_c\in\mathbb{R}^{d}$ is the L2-normalized average of text features across a bank of prompt templates; we stack them into $\mathbf{T}=[\mathbf{t}_1,\dots,\mathbf{t}_C]^\top\in\mathbb{R}^{C\times d}$. Global logits (pre-softmax) are $\mathbf{z}(\mathbf{x})=\tau\,\mathbf{T}\,\mathbf{v}(\mathbf{x})\in\mathbb{R}^{C}$, where $\tau>0$ is the fixed logit scale of the VLM. The overall architecture of OGKD framework is illustrated in Fig.~\ref{fig:detail-method-wide}.            

\subsection{Class-Similarity Graph}
\label{sec:graph_v2}
Our prior is that semantically related classes (\eg \emph{benign} vs.\ \emph{malignant}) should be closer than unrelated ones. We encode this relation in a fixed class-similarity matrix \(\mathbf{W}\in\mathbb{R}^{C\times C}\) computed once from the frozen text encoder and kept constant during training. Each row \(\mathbf{W}_{c:}\) contains non-negative weights over \(\{1,\ldots,C\}\) that sum to one, quantifying the nearest neighbors of class \(c\).

\para{Text prototypes}
For each class \(c \in \{1,\dots,C\}\), we obtain a textual prototype \(\mathbf{t}_c \in \mathbb{R}^d\) by averaging text features computed with a \emph{frozen} text encoder over a bank of prompt templates (\eg \(50\) per class), followed by L2 normalization. Stacking them row-wise gives \(\mathbf{T}=[\mathbf{t}_1,\dots,\mathbf{t}_C]^\top \in \mathbb{R}^{C \times d}\), whose \(c\)-th row equals \(\mathbf{t}_c^\top\).

\para{Pairwise similarities}
We quantify semantic relations between classes \(i\) and \(j\) using cosine similarity
\begin{equation}
\begin{aligned}
  s_{ij} &= \cos\!\big(\mathbf{t}_i,\mathbf{t}_j\big) = \mathbf{t}_i^\top \mathbf{t}_j \in [-1,1],\\
  \mathbf{s}_{c:} &= [\,s_{c1},\dots,s_{cC}\,] \in \mathbb{R}^{C}.
\end{aligned}
\label{eq:pairwise-cos}
\end{equation}
We then convert each similarity row $\mathbf{s}_{c:}$ into a probability-like distribution with a row-wise softmax:
\begin{equation}
  \mathbf{W}_{c:} = \operatorname{softmax}\!\big(\alpha\mathbf{s}_{c:}\big) \in \mathbb{R}^{C},
  \qquad \mathbf{W}_{cd}\ge 0,\quad \sum_{d=1}^{C}\mathbf{W}_{cd}=1.
  \label{eq:row-softmax}
\end{equation}
Stacking all rows gives \(\mathbf{W}=\begin{bmatrix}\mathbf{W}_{1:}\\[-2pt]\vdots\\[-2pt]\mathbf{W}_{C:}\end{bmatrix}\in\mathbb{R}^{C\times C}\), where \(\alpha>0\) controls graph sharpness: larger \(\alpha\) yields a sharper neighborhood; smaller \(\alpha\) spreads weight more broadly. \(\mathbf{W}\) depends only on frozen, text-driven prototypes and a fixed teacher encoder-not on any training images to avoid overfitting in few-shot settings.

\subsection{Global Geometry-Aware Distillation}
\label{sec:gakl-global}
We shape the teacher signal so that the probability mass does not flow uniformly to all non-targets but is redistributed to \emph{semantically related} classes according to the graph \(\mathbf{W}\) from Sec.~\ref{sec:graph_v2}. This yields a softer yet \emph{directional} target that reduces spurious confidence on distant classes.

\para{Student and teacher logits}
Let \(\hat{\mathbf{T}}\in\mathbb{R}^{C\times d}\) denote the student textual features obtained from frozen text encoder with the learnable context \(\mathbf{C}\), and let \(\tilde{\mathbf{T}}\in\mathbb{R}^{C\times d}\) denote the \emph{teacher} textual prototypes formed by averaging a subset of LLM prompts retained by a statistics-based selector per class from BiomedCoOp~\citep{koleilat2025biomedcoop}. Given an image \(\mathbf{x}\) with global image token \(\mathbf{v}(\mathbf{x})\in\mathbb{R}^{d}\) and fixed logit scale \(\tau>0\), the student and teacher global logits are:
\begin{equation}
  \mathbf{z}_{\mathrm{stu}}(\mathbf{x}) = \tau\,\hat{\mathbf{T}}\,\mathbf{v}(\mathbf{x}) \in\mathbb{R}^{C},\;
  \tilde{\mathbf{z}}(\mathbf{x}) = \tau\,\tilde{\mathbf{T}}\,\mathbf{v}(\mathbf{x}) \in\mathbb{R}^{C}.
  \label{eq:global-logits-stu-tea}
\end{equation}
Then we define the student and teacher log-probabilities with temperature \(T\)
\begin{equation}
\begin{aligned}
  \boldsymbol{\ell}_{\mathrm{stu}}(\mathbf{x}) &\,=\, \log \operatorname{softmax}\!\big(T^{-1}\,\mathbf{z}_{\mathrm{stu}}(\mathbf{x})\big),\\
  \boldsymbol{\ell}_{\mathrm{tea}}(\mathbf{x}) &\,=\, \log \operatorname{softmax}\!\big(T^{-1}\,\tilde{\mathbf{z}}(\mathbf{x})\big).
\end{aligned}
\label{eq:log-softmax-stu-tea}
\end{equation}
To inject class geometry, we diffuse the teacher with graph \(\mathbf{W}\) using a hyperparameter \(\gamma\in[0,1]\) that controls the geometry-aware strength:
\begin{equation}
  \boldsymbol{\ell}_{\mathrm{tea}}^{\star}(\mathbf{x}) = (1-\gamma)\,\boldsymbol{\ell}_{\mathrm{tea}}(\mathbf{x}) + \gamma\,\boldsymbol{\ell}_{\mathrm{tea}}(\mathbf{x})\,\mathbf{W},
  \label{eq:log-smoothing}
\end{equation}
which partially reallocates teacher knowledge toward graph neighbors of ground-truth class. Given a mini-batch \(\mathcal{B}\), the Global Geometry--Aware Distillation loss (GAD) is:
\begin{equation}
  \mathcal{L}_{\mathrm{GAD}} = \frac{1}{|\mathcal{B}|} \sum_{\mathbf{x}\in\mathcal{B}} \mathrm{KL}\!\left( \boldsymbol{\ell}_{\mathrm{tea}}^{\star}(\mathbf{x}) \,\middle\|\, \boldsymbol{\ell}_{\mathrm{stu}}(\mathbf{x}) \right),
  \label{eq:gakl}
\end{equation}
where \(\mathrm{KL}(\cdot\|\cdot)\) denotes the KL divergence with \emph{log-targets}. \(\mathcal{L}_{\mathrm{GAD}}\) encourages the student to remain confident on ground truth while considering \emph{semantically nearby} classes, reducing spurious confidence in unrelated alternatives.

\subsection{Label-Guided Geometry Distillation}
\label{sec:omni-attn}
Beyond global \IMG{} token, patch tokens contain fine-grained evidence that is valuable in biomedical images. We therefore distill at the level of \emph{patch tokens} while injecting the same class geometry used in \cref{sec:gakl-global}. This produces a supervision signal that is both attentive (focused on the most informative tokens) and geometry-aware.

\para{Patch tokens and logits}
Let the frozen image encoder produce \(P\) patch embeddings for \(\mathbf{x}\), projected to the shared space as \(\mathbf{Z}(\mathbf{x})=[\mathbf{z}_1(\mathbf{x}),\dots,\mathbf{z}_P(\mathbf{x})]^\top \in \mathbb{R}^{P\times d}\). Given the student text embedding \(\hat{\mathbf{T}}\in\mathbb{R}^{C\times d}\) and teacher text embedding \(\tilde{\mathbf{T}}\in\mathbb{R}^{C\times d}\) (see \cref{sec:gakl-global}), the student and teacher \emph{patch--class} logits are
\begin{equation}
\begin{aligned}
  \mathbf{S}(\mathbf{x}) = \tau\,\mathbf{Z}(\mathbf{x})\,\hat{\mathbf{T}}^{\top} \in \mathbb{R}^{P\times C},\\ 
  \tilde{\mathbf{S}}(\mathbf{x}) = \tau\,\mathbf{Z}(\mathbf{x})\,\tilde{\mathbf{T}}^{\top} \in \mathbb{R}^{P\times C}.
\end{aligned}
\label{eq:patch-logits}
\end{equation}
We distill the student using informative patch tokens, supervised on regions aligned with ground-truth signal. For each \((\mathbf{x},y)\), we score patch token \(n\) by its \emph{alignment} with teacher text embedding for \emph{ground-truth} class:
\begin{equation}
  s_n(\mathbf{x},y) = \cos\!\big(\mathbf{z}_n(\mathbf{x}),\,\tilde{\mathbf{t}}_y\big) = \frac{\mathbf{z}_n(\mathbf{x})^\top \tilde{\mathbf{t}}_y}{\|\mathbf{z}_n(\mathbf{x})\|_2\,\|\tilde{\mathbf{t}}_y\|_2}.
  \label{eq:cos-score}
\end{equation}
We then retain the \(K\) most attentive patches (highest \(s_n\)):
\begin{equation}
  \mathcal{I}_K(\mathbf{x},y) = \operatorname{TopK}_K\!\big(\{\,s_n(\mathbf{x},y)\,\}_{n=1}^{P}\big).
  \label{eq:topk-threshold}
\end{equation}
(For example, \(K=\lfloor 0.10\,P\rfloor\) retains the top \(10\%\).) Restricting \(\mathbf{S}(\mathbf{x})\) and \(\tilde{\mathbf{S}}(\mathbf{x})\) to the selected patch tokens reduces their shapes from \(\mathbb{R}^{P\times C}\) to \(\mathbb{R}^{K\times C}\). For the selected patches, we form log-probabilities over classes,
\begin{equation}
\begin{aligned}
  \mathbf{L}^{\mathrm{stu}}(\mathbf{x}) = \log \operatorname{softmax}\!\big(T^{-1}\,\mathbf{S}(\mathbf{x})\big),\\
  \mathbf{L}^{\mathrm{tea}}(\mathbf{x}) = \log \operatorname{softmax}\!\big(T^{-1}\,\tilde{\mathbf{S}}(\mathbf{x})\big).
\end{aligned}
\label{eq:patch-logps}
\end{equation}
and diffuse the teacher along the graph \(\mathbf{W}\):
\begin{equation}
  \mathbf{L}^{\mathrm{tea}\star}(\mathbf{x}) = (1-\gamma)\,\mathbf{L}^{\mathrm{tea}}(\mathbf{x}) + \gamma\,\mathbf{L}^{\mathrm{tea}}(\mathbf{x})\,\mathbf{W}.
  \label{eq:patch-graph-smooth}
\end{equation}
Let \(\mathbf{e}_y\in\{0,1\}^{C}\) be the one-hot vector for class \(y\). Gathering the \emph{ground-truth} coordinate and restricting to \(\mathcal{I}_K(\mathbf{x},y)\) gives
\begin{equation}
\begin{aligned}
  \mathbf{u}^{\mathrm{stu}}_{\mathcal{I}}(\mathbf{x},y) &= \big(\mathbf{L}^{\mathrm{stu}}(\mathbf{x})\,\mathbf{e}_y\big)\Big|_{\mathcal{I}_K(\mathbf{x},y)} \in \mathbb{R}^{K},\\
  \mathbf{u}^{\mathrm{tea}\star}_{\mathcal{I}}(\mathbf{x},y) &= \big(\mathbf{L}^{\mathrm{tea}\star}(\mathbf{x})\,\mathbf{e}_y\big)\Big|_{\mathcal{I}_K(\mathbf{x},y)} \in \mathbb{R}^{K}.
\end{aligned}
\label{eq:gather-selected}
\end{equation}
The Label-Guided Geometry Distillation (LGD) loss is
\begin{equation}
  \mathcal{L}_{\mathrm{LGD}} = \frac{1}{|\mathcal{B}|}\sum_{(\mathbf{x},y)\in\mathcal{B}}
  \operatorname{KL}\!\left(\mathbf{u}^{\mathrm{tea}\star}_{\mathcal{I}}(\mathbf{x},y) \,\middle\|\, \mathbf{u}^{\mathrm{stu}}_{\mathcal{I}}(\mathbf{x},y)\right).
  \label{eq:lgd}
\end{equation}

\para{Discussion}
LGD introduces no trainable parameters and reuses frozen patch tokens and text prototypes. While GAD (\cref{sec:gakl-global}) distills the \emph{full} class distribution at the global \IMG{} token, LGD is deliberately \emph{label-focused}: after making the teacher geometry-aware, we gather only the ground-truth channel at the selected patches (see \cref{eq:gather-selected}) to emphasize fine-grained, label-specific evidence and reduce compute by avoiding all patch tokens.

\subsection{Overall Training Objective}
\textbf{Classification loss.}
Given student logits \(\mathbf{z}_{\mathrm{stu}}(\mathbf{x})=\tau\,\hat{\mathbf{T}}\,\mathbf{v}(\mathbf{x})\in\mathbb{R}^{C}\) from \cref{eq:global-logits-stu-tea}, the predicted probabilities are \(\mathbf{p}_{\mathrm{stu}}(\mathbf{x})=\operatorname{softmax}\big(\mathbf{z}_{\mathrm{stu}}(\mathbf{x})\big)\).
For a mini-batch \(\mathcal{B}\) with labels \(y\in\{1,\dots,C\}\), we use conventional cross-entropy loss for the classification task:
\begin{equation}
  \mathcal{L}_{\mathrm{CE}} = \frac{1}{|\mathcal{B}|} \sum_{(\mathbf{x},y)\in\mathcal{B}} \Bigl[-\log\bigl(\mathbf{p}_{\mathrm{stu}}(\mathbf{x})\bigr)_{y}\Bigr].
  \label{eq:ce}
\end{equation}
\textbf{Semantic Consistency by Contextual Mapping (SCCM).}
This SCCM loss from BiomedCoOp~\citep{koleilat2025biomedcoop} is defined as:
Let \(\hat{\mathbf{T}}=[\hat{\mathbf{t}}_1,\dots,\hat{\mathbf{t}}_C]^\top\in\mathbb{R}^{C\times d}\) denote the \emph{student} text features produced by the frozen text encoder with the learnable context \(\mathbf{C}\), and let \(\mathbf{T}=[\mathbf{t}_1,\dots,\mathbf{t}_C]^\top\in\mathbb{R}^{C\times d}\) be the \emph{frozen} per-class textual prototypes obtained by averaging zero-shot prompt templates (\cref{sec:graph_v2}). SCCM encourages the learned context to remain semantically anchored to the frozen prototypes via a per-class squared error:
\begin{equation}
  \mathcal{L}_{\mathrm{SCCM}} = \frac{1}{C}\sum_{c=1}^{C} \bigl\|\hat{\mathbf{t}}_c-\mathbf{t}_c\bigr\|_2^2 = \frac{1}{C}\,\bigl\|\hat{\mathbf{T}}-\mathbf{T}\bigr\|_{F}^{2},
  \label{eq:sccm}
\end{equation}
where \(\|\cdot\|_{F}\) denotes the Frobenius norm.

\label{sec:overall-loss}
\paragraph{The total training loss}
\begin{equation}
  \mathcal{L} = \mathcal{L}_{\mathrm{CE}} + \lambda_{1}\,\mathcal{L}_{\mathrm{SCCM}} + \lambda_{2}\,\mathcal{L}_{\mathrm{GAD}} + \lambda_{3}\,\mathcal{L}_{\mathrm{LGD}},
  \label{eq:total-loss}
\end{equation}
where \(\lambda_{1},\lambda_{2},\lambda_{3} > 0\) are weights of the respective loss.
\begin{table}[tb]
  \centering
  \caption{\textbf{Accuracy comparison on base-to-novel generalization with state-of-the-art prompt-learning methods (16-shot).} Average accuracy (\%) over 3 runs. Best per column is in \textbf{bold}; second best is \underline{underlined}. HM denotes the harmonic mean between base and novel. Citations are provided in the ``AVERAGE over 10 DATASETS'' table.}
  \label{tab:base2novel_3x}
  \begingroup
  \small
  \setlength{\tabcolsep}{2.2pt}
  \renewcommand{\arraystretch}{0.95}
  \begin{minipage}[t]{0.62\linewidth}\centering
    \parbox{\linewidth}{\centering\textcolor{mygreen}{\textbf{AVERAGE over 10 DATASETS}}}\vspace{1.2pt}
    \begin{sc}
    \begin{tabular}{l@{\hspace{0.4em}} r@{\hspace{1.9em}} r@{\hspace{2.2em}} r}
      \toprule
      Method & Base & Novel & HM \\
      \midrule
      BiomedCLIP~\citep{zhang2023biomedclip}     & 47.84 & 65.42 & 53.81 \\
      CoOp~\citep{zhou2022learning}              & 73.85 & 64.75 & 67.23 \\
      CoCoOp~\citep{zhou2022conditional}         & 72.26 & 67.03 & 67.22 \\
      KgCoOp~\citep{yao2023visual}               & 68.36 & 64.08 & 64.61 \\
      ProGrad~\citep{zhu2023prompt}              & 71.67 & 66.93 & 67.43 \\
      BiomedCoOp~\citep{koleilat2025biomedcoop}  & 76.26 & 73.92 & 75.07 \\
      Ours                                     & \textbf{79.03}\upgain{2.77}
                                               & \textbf{75.60}\upgain{1.68}
                                               & \textbf{77.28}\upgain{2.21} \\
      \bottomrule
    \end{tabular}
    \end{sc}
  \end{minipage}\hfill
  \begin{minipage}[t]{0.33\linewidth}\centering
    \parbox{\linewidth}{\centering\textbf{BTMRI}}\vspace{1.2pt}
    \begin{sc}
    \begin{tabular}{lrrr}
      \toprule
      Method & Base & Novel & HM \\
      \midrule
      BiomedCLIP     & 40.88 & \underline{96.18} & 57.37 \\
      CoOp          & 82.25 & 94.51 & 87.95 \\
      CoCoOp        & 77.88 & 94.84 & 85.53 \\
      KgCoOp        & 78.03 & 95.05 & 85.69 \\
      ProGrad       & 82.13 & 94.98 & 88.09 \\
      BiomedCoOp    & \underline{82.42} & \textbf{96.84} & \underline{89.05} \\
      Ours          & \textbf{84.88} & 96.11 & \textbf{90.15} \\
      \bottomrule
    \end{tabular}
    \end{sc}
  \end{minipage}
  \vspace{1.2pt}
  \newcommand{\dsmini}[1]{\begin{minipage}[t]{0.32\linewidth}\centering
    \setlength{\tabcolsep}{1.5pt}%
    \parbox{\linewidth}{\centering\textbf{#1}}\vspace{1.2pt}}
  \newcommand{\dsminiend}{\end{minipage}}
  \dsmini{COVID-QU-Ex}
    \begin{sc}\begin{tabular}{lrrr}
      \toprule Method & Base & Novel & HM \\ \midrule
      BiomedCLIP  & 53.96 & 89.43 & 67.31 \\
      CoOp       & 75.92 & 90.07 & 82.39 \\
      CoCoOp     & \underline{77.28} & 87.61 & 82.12 \\
      KgCoOp     & 75.42 & 89.61 & 81.90 \\
      ProGrad    & 75.19 & 90.34 & 82.07 \\
      BiomedCoOp & 75.91 & \underline{91.63} & \underline{83.03} \\
      Ours       & \textbf{78.86} & \textbf{93.07} & \textbf{85.38} \\
      \bottomrule
    \end{tabular}\end{sc}
  \dsminiend\hfill
  \dsmini{CHMNIST}
    \begin{sc}\begin{tabular}{lrrr}
      \toprule Method & Base & Novel & HM \\ \midrule
      BiomedCLIP  & 37.63 & 40.69 & 39.10 \\
      CoOp       & \underline{89.41} & 35.11 & 50.42 \\
      CoCoOp     & 87.77 & 42.51 & 57.28 \\
      KgCoOp     & 75.45 & 38.70 & 51.16 \\
      ProGrad    & 82.98 & \underline{44.19} & 57.67 \\
      BiomedCoOp & 88.87 & 42.73 & \underline{57.71} \\
      Ours       & \textbf{89.54} & \textbf{49.82} & \textbf{64.02} \\
      \bottomrule
    \end{tabular}\end{sc}
  \dsminiend\hfill
  \dsmini{DermaMNIST}
    \begin{sc}\begin{tabular}{lrrr}
      \toprule Method & Base & Novel & HM \\ \midrule
      BiomedCLIP  & 34.95 & 49.59 & 41.00 \\
      CoOp       & 48.06 & 59.41 & 53.14 \\
      CoCoOp     & 42.88 & 60.66 & 50.24 \\
      KgCoOp     & 36.41 & 47.31 & 41.15 \\
      ProGrad    & 35.52 & \underline{63.28} & 45.50 \\
      BiomedCoOp & \underline{54.86} & \textbf{74.10} & \underline{63.04} \\
      Ours       & \textbf{65.45} & \underline{67.31} & \textbf{66.37} \\
      \bottomrule
    \end{tabular}\end{sc}
  \dsminiend
  \vspace{1.2pt}
  \dsmini{Kvasir}
    \begin{sc}\begin{tabular}{lrrr}
      \toprule Method & Base & Novel & HM \\ \midrule
      BiomedCLIP  & 75.00 & 60.50 & 66.97 \\
      CoOp       & 86.22 & 58.06 & 69.39 \\
      CoCoOp     & 85.94 & 53.95 & 66.29 \\
      KgCoOp     & 81.56 & 59.00 & 68.47 \\
      ProGrad    & 82.89 & 60.45 & 69.91 \\
      BiomedCoOp & \underline{86.50} & \underline{61.83} & \underline{72.11} \\
      Ours       & \textbf{86.83} & \textbf{66.61} & \textbf{75.39} \\
      \bottomrule
    \end{tabular}\end{sc}
  \dsminiend\hfill
  \dsmini{CTKIDNEY}
    \begin{sc}\begin{tabular}{lrrr}
      \toprule Method & Base & Novel & HM \\ \midrule
      BiomedCLIP  & 38.55 & 52.99 & 44.63 \\
      CoOp       & 82.24 & 67.92 & 74.40 \\
      CoCoOp     & 81.96 & 56.56 & 66.93 \\
      KgCoOp     & 81.67 & 58.45 & 68.14 \\
      ProGrad    & 83.86 & 63.01 & 71.96 \\
      BiomedCoOp & \textbf{86.93} & \underline{78.94} & \textbf{82.74} \\
      Ours       & \underline{83.88} & \textbf{81.11} & \underline{82.47} \\
      \bottomrule
    \end{tabular}\end{sc}
  \dsminiend\hfill
  \dsmini{LC25000}
    \begin{sc}\begin{tabular}{lrrr}
      \toprule Method & Base & Novel & HM \\ \midrule
      BiomedCLIP  & 59.73 & 87.60 & 71.03 \\
      CoOp       & 90.12 & 87.55 & 88.82 \\
      CoCoOp     & 88.33 & 95.02 & 91.55 \\
      KgCoOp     & 88.13 & 86.44 & 87.28 \\
      ProGrad    & 90.29 & 85.47 & 87.81 \\
      BiomedCoOp & \underline{93.77} & \underline{97.00} & \underline{95.36} \\
      Ours       & \textbf{94.34} & \textbf{97.98} & \textbf{96.13} \\
      \bottomrule
    \end{tabular}\end{sc}
  \dsminiend
  \vspace{1.2pt}
  \dsmini{RETINA}
    \begin{sc}\begin{tabular}{lrrr}
      \toprule Method & Base & Novel & HM \\ \midrule
      BiomedCLIP  & 45.18 & 55.28 & 49.72 \\
      CoOp       & \underline{70.98} & 56.90 & 63.16 \\
      CoCoOp     & 66.88 & 65.56 & 66.21 \\
      KgCoOp     & 60.77 & 54.91 & 57.69 \\
      ProGrad    & 68.77 & 58.43 & 63.18 \\
      BiomedCoOp & 68.46 & \underline{67.72} & \underline{68.09} \\
      Ours       & \textbf{75.57} & \textbf{73.70} & \textbf{74.62} \\
      \bottomrule
    \end{tabular}\end{sc}
  \dsminiend\hfill
  \dsmini{KneeXray}
    \begin{sc}\begin{tabular}{lrrr}
      \toprule Method & Base & Novel & HM \\ \midrule
      BiomedCLIP  & 35.89 & 71.90 & 47.88 \\
      CoOp       & 38.28 & 47.69 & 42.47 \\
      CoCoOp     & 34.08 & 63.14 & 44.27 \\
      KgCoOp     & 37.94 & 61.19 & 46.84 \\
      ProGrad    & 40.88 & 59.12 & 48.34 \\
      BiomedCoOp & \underline{44.23} & \textbf{78.35} & \underline{56.54} \\
      Ours       & \textbf{45.56} & \underline{77.49} & \textbf{57.38} \\
      \bottomrule
    \end{tabular}\end{sc}
  \dsminiend\hfill
  \dsmini{OCTMNIST}
    \begin{sc}\begin{tabular}{lrrr}
      \toprule Method & Base & Novel & HM \\ \midrule
      BiomedCLIP  & 56.60 & 50.00 & 53.10 \\
      CoOp       & 75.00 & 50.23 & 60.17 \\
      CoCoOp     & 79.60 & \underline{50.47} & \underline{61.77} \\
      KgCoOp     & 68.20 & 50.13 & 57.79 \\
      ProGrad    & 74.20 & 50.02 & 59.76 \\
      BiomedCoOp & \underline{80.33} & 50.07 & 61.69 \\
      Ours       & \textbf{85.40} & \textbf{52.80} & \textbf{65.25} \\
      \bottomrule
    \end{tabular}\end{sc}
  \dsminiend
  \endgroup
\end{table}
\newcommand{\NA}{\multicolumn{1}{c}{---}}
\newcommand{\mstd}{\,\(\pm\)\,}
\begin{table}[t]
\caption{\textbf{Few-shot evaluation against state-of-the-art methods.} Average classification accuracy (\%) over 11 datasets, reported as mean\(\pm\)std over 3 runs. Best results for each \(K\) (shots) are in \textbf{bold}.}
\label{tab:fewshot_sota}
\centering
\begin{small}
\setlength{\tabcolsep}{3pt}
\renewcommand{\arraystretch}{1.08}
\resizebox{\linewidth}{!}{%
\begin{tabular}{lccccc}
\toprule
Method & \(K=1\) & \(K=2\) & \(K=4\) & \(K=8\) & \(K=16\) \\
\midrule
\multicolumn{6}{l}{\emph{Zero-shot methods}}\\
BiomedCLIP~\citep{zhang2023biomedclip}                      & \NA & \NA & 42.05 & \NA & \NA \\
BiomedCLIP~\citep{zhang2023biomedclip} + Ensemble           & \NA & \NA & 52.27 & \NA & \NA \\
BiomedCLIP~\citep{zhang2023biomedclip} + Selective Ensemble & \NA & \NA & 53.72 & \NA & \NA \\
\midrule
\multicolumn{6}{l}{\emph{CLIP-based adapter methods}}\\
CLIP-Adapter~\citep{gao2024clip}   & 44.66\mstd 2.97 & 43.91\mstd 2.48 & 44.36\mstd 1.94 & 45.42\mstd 2.38 & 46.69\mstd 1.71 \\
Tip-Adapter~\citep{zhang2021tip}   & 49.19\mstd 4.84 & 52.36\mstd 6.57 & 57.33\mstd 5.07 & 61.98\mstd 5.76 & 67.15\mstd 4.25 \\
Tip-Adapter-F~\citep{zhang2021tip} & 51.17\mstd 8.33 & 52.74\mstd 5.88 & 61.23\mstd 6.22 & 65.91\mstd 3.64 & 70.91\mstd 2.65 \\
\midrule
\multicolumn{6}{l}{\emph{Linear probing methods}}\\
LP++~\citep{huang2024lp++}              & 47.24\mstd 7.68 & 53.18\mstd 7.29 & 59.02\mstd 6.93 & 63.69\mstd 4.68 & 68.35\mstd 3.59 \\
\midrule
\multicolumn{6}{l}{\emph{Prompt learning methods}}\\
CoOp~\citep{zhou2022learning}              & 50.16\mstd 6.93 & 54.18\mstd 4.31 & 59.75\mstd 3.72 & 65.84\mstd 3.66 & 69.62\mstd 2.83 \\
CoCoOp~\citep{zhou2022conditional}         & 48.49\mstd 4.39 & 51.28\mstd 5.06 & 54.69\mstd 4.79 & 61.08\mstd 3.49 & 65.09\mstd 2.87 \\
KgCoOp~\citep{yao2023visual}               & 50.85\mstd 5.59 & 53.18\mstd 4.33 & 57.82\mstd 4.50 & 62.08\mstd 2.59 & 62.84\mstd 1.72 \\
ProGrad~\citep{zhu2023prompt}              & 51.88\mstd 6.39 & 54.71\mstd 4.46 & 60.42\mstd 4.78 & 65.61\mstd 3.02 & 67.13\mstd 3.00 \\
BiomedCoOp~\citep{koleilat2025biomedcoop}  & 57.03\mstd 2.80 & 59.13\mstd 3.64 & 63.95\mstd 2.42 & 68.32\mstd 2.65 & 72.42\mstd 1.69 \\
\textbf{Ours}                             & \textbf{59.40\mstd 2.86} & \textbf{61.10\mstd 2.08} & \textbf{65.32\mstd 2.26} & \textbf{70.84\mstd 1.81} & \textbf{73.80\mstd 1.38} \\
\bottomrule
\end{tabular}%
}
\end{small}
\end{table}

\section{Experiments and Analysis}
\subsection{Experimental setup and results}
\label{sec:exp-setup}

\para{Protocols}
We evaluate our approach under two few-shot recognition protocols. (i)~\emph{Few-shot learning:} for each dataset, we sample $K\!\in\!\{1,2,4,8,16\}$ labeled images per class to form a support set, and train only the prompt learner while keeping the vision--language backbone frozen. Performance is reported on the test split. (ii)~\emph{Base--to--novel generalization:} each dataset is divided into base and novel classes. The model is trained on the base classes with \(K=16\) shots and then evaluated on both base and novel test sets, reporting base accuracy, novel accuracy, and harmonic mean (HM) to quantify transfer to unseen categories. We also evaluate the robustness of our method and BiomedCoOp~\citep{koleilat2025biomedcoop} under label-noise few-shot conditions (details are provided in \cref{app:label-noise}).

\para{Datasets}
We conduct experiments on 11 biomedical image datasets spanning nine imaging modalities and ten anatomical sites: CTKIDNEY~\citep{islam2022vision} (CT), DermaMNIST~\citep{codella2019skin, tschandl2018ham10000} (dermatoscopy), Kvasir~\citep{pogorelov2017kvasir} (endoscopy), RETINA~\citep{kohler2013automatic, porwal2018indian} (fundus photography), LC25000~\citep{borkowski2019lung} and CHMNIST~\citep{kather2016multi} (histopathology), BTMRI~\citep{nickparvar2021brain} (brain MRI), OCTMNIST~\citep{kermany2018identifying} (OCT), BUSI~\citep{al2020dataset} (ultrasound), COVID\mbox{-}QU\mbox{-}Ex~\citep{tahir2021covid} and KneeXray~\citep{chen2018knee} (X-ray). We follow the official train/validation/test splits for each dataset. For the base--to--novel benchmark, BUSI~\citep{al2020dataset} is excluded due to limited class diversity; it is still included in few-shot evaluation.

\para{Implementation details}
We use BiomedCLIP~\citep{zhang2023biomedclip} (ViT-B/16) as the vision--language backbone. Only learnable context tokens are optimized; all encoder parameters remain frozen. For base--to--novel generalization, we train for 50 epochs; for few-shot evaluation, we train for 100 epochs. We use SGD with learning rate $2.5\times 10^{-3}$, batch size 4, and random resized crop for data augmentation. We initialize the learnable context with ``a photo of a'', and set the LLM prompt bank to 50 templates per class when building teacher text prototypes. We keep the backbone and training recipe consistent with BiomedCoOp~\citep{koleilat2025biomedcoop} to ensure a fair comparison. The geometry strength \(\gamma\), top-token ratio \(k\%\), \(\lambda_{2}\), \(\lambda_{3}\) for each dataset are detailed in the appendix document. For \(\lambda_{1}\), we use the same weight from BiomedCoOp. We set graph sharpness \(\alpha\) to 1 in base--to--novel generalization and 4 in few-shot evaluation. All main and ablation results in this work are averaged over 3 independent runs. All experiments were run on a single NVIDIA A100 GPU (40\,GB). Additional hyperparameters and pseudocode are provided in \cref{app:hyperparameters_pseudocode}.

\para{Base--to--novel generalization results}
As shown in Table~\ref{tab:base2novel_3x}, our method outperforms BiomedCoOp~\citep{koleilat2025biomedcoop} by $(+2.77/1.68/2.21)\%$ (Base/Novel/HM) on average across 10 datasets. It achieves the best HM on 9/10 datasets, indicating strong transfer to novel classes. In particular, compared to BiomedCoOp, we observe large gains on RETINA $(+7.11/5.98/6.53)\%$, CHMNIST $(+0.67/7.09/6.31)\%$, OCTMNIST $(+5.07/2.73/3.56)\%$, and COVID\mbox{-}QU\mbox{-}Ex $(+2.95/1.44/2.35)\%$. These results highlight the benefit of geometry-aware distillation across diverse imaging modalities.

\para{Few-shot evaluation results}
As shown in Table~\ref{tab:fewshot_sota}, our method achieves strong accuracy and consistently improves over BiomedCoOp~\citep{koleilat2025biomedcoop} by about 2\% on average across \(K\in\{1,2,4,8,16\}\). In particular, at \(K=1\) and \(K=8\), we outperform BiomedCoOp by about 2.4\%, demonstrating the effectiveness of the method.

\subsection{Method analysis}
\begin{figure}[t]
  \centering
  \includegraphics[width=0.7\linewidth]{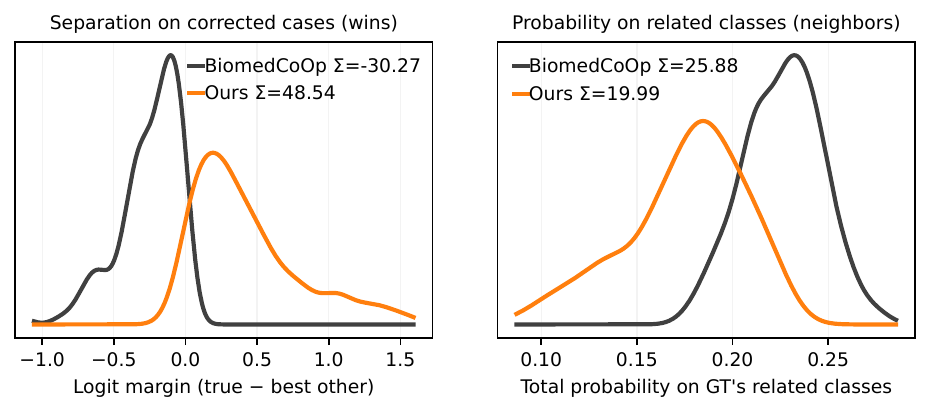}
  \caption{\textbf{Wins analysis via KDE curves} on the RETINA dataset. We compare BiomedCoOp and our method on challenging test samples that our method classifies correctly while BiomedCoOp is wrong (``wins''). \emph{Left:} KDE of the logit margin \(z(y)-\max_{j\neq y} z(j)\). \emph{Right:} KDE of the probability assigned to the ground-truth class's top-\(k\) neighbors in \(W\) (here \(k{=}3\)). The legend in each panel also reports \(\Sigma\), the sum over all win samples.}
  \label{fig:wins-kde-two-panel}
  \vspace{-0.15in}
\end{figure}

\para{Why geometry-aware design helps}
On samples that our method classifies correctly but BiomedCoOp~\citep{koleilat2025biomedcoop} misclassifies, the margin distribution (as shown in Fig.~\ref{fig:wins-kde-two-panel}, left panel) shifts right and exhibits a larger \(\Sigma\), indicating a stronger separation between the true class and its closest neighbors. In the right panel, BiomedCoOp~\citep{koleilat2025biomedcoop} allocates more probability to \emph{neighbors} of the ground-truth class (larger \(\Sigma\) on the neighbor-sum curve), revealing near-miss behavior. Our geometry-aware design leverages the neighborhood structure to discourage spreading mass onto related but non-target classes, which sharpens the target logit and increases the margin. These two diagnostics: fewer near-misses neighbor and larger decision margins explain the observed gains in few-shot accuracy.

\begin{figure}[t]
  \centering
  \includegraphics[width=0.7\linewidth]{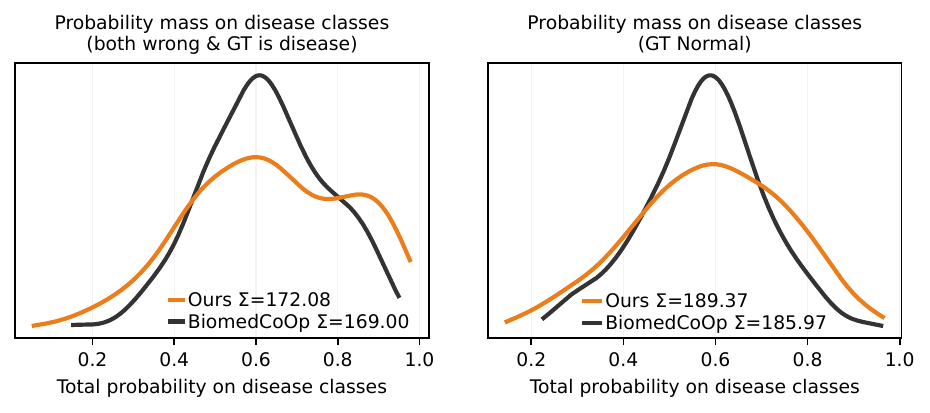}
  \caption{\textbf{Probability mass on disease classes (excluded GT)} on Retina dataset. Each curve is a kernel density estimate over per-image quantity \(s(x)=\sum_{c\in\mathcal{D}} p_\theta(y=c \mid x)\), where \(\mathcal{D}\) are the disease classes (Normal excluded). Legends report \(\sum_x s(x)\) (\(\Sigma\)): total probability mass placed on disease classes across all images in the subset. \textbf{Left:} subset where both methods misclassify and the ground-truth class is a disease. \textbf{Right:} all images with normal GT.}
  \label{fig:kde-disease-mass-balanced}
\end{figure}

\begin{figure}[t]
  \centering
  \includegraphics[width=\textwidth]{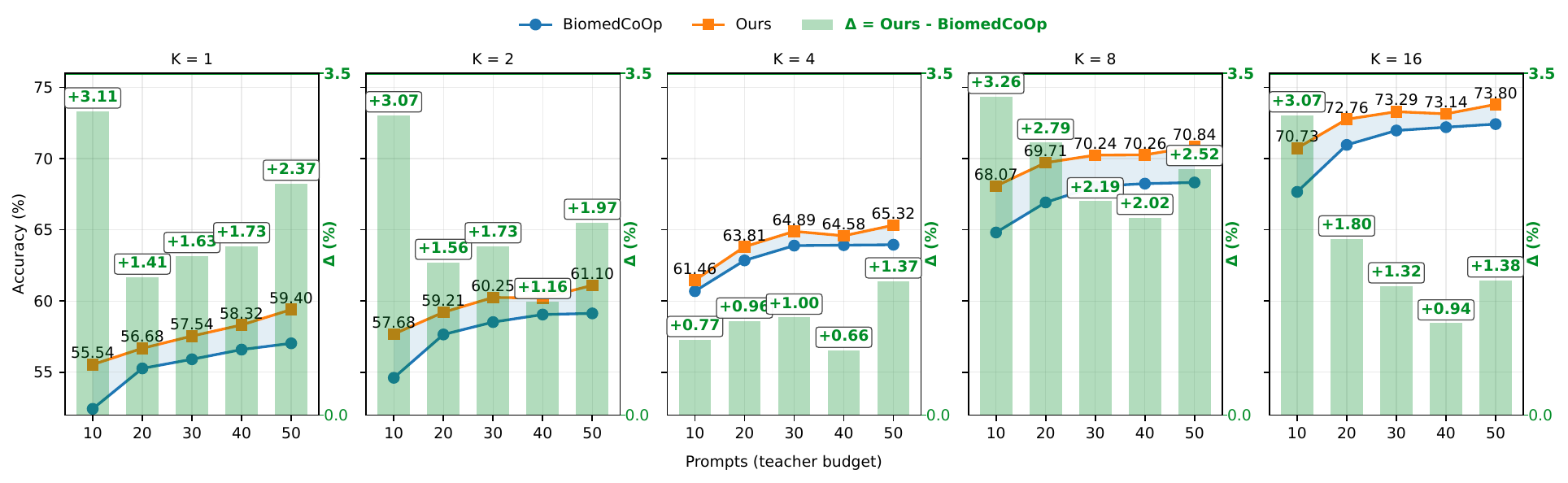}%
  \caption{\textbf{Few-shot evaluation averaged over 11 datasets under different teacher budgets} (number of LLM prompt templates per class) with shots \(K\in\{1,2,4,8,16\}\). The left y-axis shows absolute accuracies (\%) for BiomedCoOp~\citep{koleilat2025biomedcoop} and our method; the right y-axis and the green strip show the per-budget improvement \(\Delta=\text{Ours}-\text{BiomedCoOp}\) (\%), annotated at each budget.}
  \label{fig:teacher-budget}
\end{figure}

\para{Analysis on misclassified samples}
We also analyze samples that \textit{both methods misclassify} and show some interesting insights. In the left panel of Fig.~\ref{fig:kde-disease-mass-balanced}, where the ground truth is a disease class, the density of our method is shifted to the right and shows a heavier right tail, and its \(\Sigma\) is larger than BiomedCoOp~\citep{koleilat2025biomedcoop}. Although both models predict the wrong \emph{specific} disease, our geometry-aware learner allocates more probability mass to \emph{disease} classes overall. If we change our perspective from multi to binary classification, through a screening lens (``disease vs.\ normal''), these are \emph{clinically safer} errors: patients would still be identified as ``likely diseased'' and sent for further examination, rather than being incorrectly diagnosed as normal. This inclination is necessary in healthcare system.

\para{Discussion} While beneficial, our method also exhibits a trade-off. In the right panel (Fig.~\ref{fig:kde-disease-mass-balanced}), we investigate the case when the ground truth is normal. The curve of our method also shifts to the right and its \(\Sigma\) increases, revealing a higher tendency to place mass on disease classes---i.e., a higher false-positive rate. This is an expected trade-off of our geometry-aware design: avoiding \emph{distant} mistakes on ambiguous disease cases can raise the probability on disease for truly normal images. We provide ideas that can help alleviate this phenomenon in the conclusion section.

\section{Ablation study and analysis}
\paragraph{Components effect.}
As shown in Table~\ref{tab:ablation_components_single}, starting from BiomedCoOp~\citep{koleilat2025biomedcoop}, replacing its vanilla KD loss with our GAD loss significantly improves both base-to-novel generalization and few-shot evaluation. Adding LGD on top yields the best performance. Overall, GAD and LGD provide consistent gains across settings.
\begin{table}[t]
  \centering
  \caption{Impact of our components. B denotes BiomedCoOp~\citep{koleilat2025biomedcoop}.}
  \label{tab:ablation_components_single}
  \small
  \begin{tabular}{@{}lccccccccc@{}}
    \toprule
    & \multicolumn{3}{c}{\textbf{Base-to-novel}} & \multicolumn{5}{c}{\textbf{Few-shot}} \\
    \cmidrule(lr){2-4} \cmidrule(lr){5-9}
    \textbf{Method} & Base & Novel & HM & $K{=}1$ & $K{=}2$ & $K{=}4$ & $K{=}8$ & $K{=}16$ \\
    \midrule
    B & 76.26 & 73.92 & 75.07 & 57.03 & 59.13 & 63.95 & 68.32 & 72.42 \\
    B + GAD & 78.61 & 74.60 & 76.56 & 57.95 & 60.39 & 64.51 & 69.99 & 73.41 \\
    B + GAD + LGD & \textbf{79.03} & \textbf{75.60} & \textbf{77.28} & \textbf{59.40} & \textbf{61.10} & \textbf{65.32} & \textbf{70.84} & \textbf{73.80} \\
    \bottomrule
  \end{tabular}
\end{table} 

\begin{table}[t]
\centering
\footnotesize
\setlength{\tabcolsep}{5.5pt}
\caption{\textbf{\(\mathcal{L}_{\mathrm{LGD}}\) variants.}
"Full (w/o geom)" disables geometry; "Full geom" distills all patch tokens with geometry; "Label-guided geom" applies geometry only at attentive tokens.}
\label{tab:ablation_attentive_tokens_all}
\resizebox{0.59\linewidth}{!}{%
\begin{tabular}{lccc}
\toprule
\textbf{Few-shot} & \textbf{Full (w/o)} & \textbf{Full geom} & \textbf{Label-guided geom} \\
\midrule
1   & 58.00 & 58.87 & \textbf{59.40} \\
2   & 60.17 & 60.70 & \textbf{61.10} \\
4   & 64.86 & 64.79 & \textbf{65.32} \\
8   & 69.88 & 70.29 & \textbf{70.84} \\
16  & 72.48 & 73.27 & \textbf{73.80} \\
\midrule
\multicolumn{4}{l}{\textbf{Base-to-novel}} \\
Base  & 78.44 & 78.49 & \textbf{79.03} \\
Novel & 73.79 & 74.38 & \textbf{75.60} \\
HM    & 76.04 & 76.38 & \textbf{77.28} \\
\bottomrule
\end{tabular}
}
\end{table}

\para{Teacher prompt budget}
We vary the teacher prompt budget across few-shot settings and report averages over 11 datasets. As shown in \cref{fig:teacher-budget}, our method consistently outperforms BiomedCoOp~\citep{koleilat2025biomedcoop} for every budget and shot. The gains are largest in the low-budget regime: at \(N{=}10\), improvements reach \(3.11\%\) (\(K{=}1\)), \(3.07\%\) (\(K{=}2\)), \(3.26\%\) (\(K{=}8\)), and \(3.07\%\) (\(K{=}16\)). These results suggest that our method yields robust gains when textual supervision is limited, while remaining beneficial at higher budgets.

\para{Geometry strength}
We vary \(\gamma\), which controls how strongly geometry is injected into the teacher.
\begin{wrapfigure}{r}{0.32\textwidth}
  \centering
  \vspace{-0.05in}
  \includegraphics[width=1.0\linewidth]{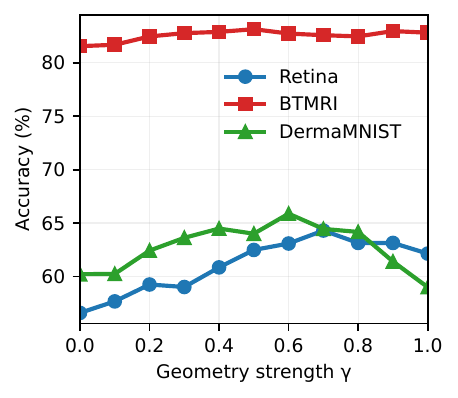}
  \vspace{-0.25in} 
  \caption{\textbf{Effect of geometry strength $\gamma$ on accuracy.}}
  \label{fig:geom-strength}
  \vspace{-0.15in}
\end{wrapfigure}
As shown in Fig.~\ref{fig:geom-strength}, without geometry (\(\gamma{=}0\)) accuracy drops substantially; moderate geometry consistently helps, whereas too much geometry (\(\gamma{=}1\)) over-regularizes class boundaries.

\para{Attentive patch tokens vs.\ full patch tokens}
In Table~\ref{tab:ablation_attentive_tokens_all}, we compare variants of \(\mathcal{L}_{\mathrm{LGD}}\) to evaluate: (i) how much the geometry-aware component improves patch-token distillation, and (ii) how label-guided alignment helps student focus on informative tokens. We observe that distilling over all patch tokens with geometry (\emph{Full geom}) outperforms the version without geometry (\emph{Full (w/o)}), demonstrating the benefit of the class-relation graph. The label-guided variant (\emph{Label-guided geom}), which distills only ground-truth channel on top-\(K\) attentive patches, further improves accuracy by concentrating supervision on regions that align strongly with ground-truth signals. Overall, focusing distillation on informative patches strengthens fine-grained evidence.

\section{Conclusion}
We introduced OGKD, a framework for prompt tuning biomedical VLMs that distills geometry-aware knowledge at both global and patch-token levels. OGKD improves classification accuracy over state-of-the-art VLM prompt-learning methods across 11 datasets and multiple evaluation settings. Using risk-coverage analysis, we further show that OGKD yields more reliable predictions, which is important for medical applications. One trade-off we observe (right panel of Fig.~\ref{fig:kde-disease-mass-balanced}) is a slight increase in false positives on \emph{Normal} class. A straightforward mitigation is to set geometry strength \(\gamma\) to zero (or a small value) for \emph{Normal}, preventing probability mass from diffusing from \emph{Normal} to its disease neighbors. This class-conditional design may reduce false positives on \emph{Normal} while retaining geometry-aware benefits for disease categories; we leave assessing this mitigation to future work.

\section*{Limitations and Ethical Statement}
\paragraph{Limitations.}
Our method is evaluated on public biomedical benchmarks, and its performance may not directly transfer to clinical deployment settings with different scanners, patient demographics, or acquisition protocols. Further validation on larger datasets collected from multiple institutions is needed before considering real-world clinical use.

\paragraph{Ethical statement.}
This work studies few-shot adaptation of biomedical vision-language models for medical image recognition. A potential positive impact is improved data-efficient adaptation for clinical imaging tasks where annotations are limited. However, the method is not intended for autonomous clinical decision-making. Final decisions should be made by qualified clinicians, and model performance may vary across institutions, devices, and patient demographics. All experiments are conducted on publicly available benchmark datasets, and we do not collect new patient data. Future deployment should include careful validation and human clinical supervision.

\section*{Acknowledgements}
This work is supported by the MBZUAI-WIS Joint Program for Artificial Intelligence Research.


\clearpage
\appendix
\section*{\Large{Appendix}}

\section{Method robustness under label noise}
\label{app:label-noise}
This ablation compares how our method and BiomedCoOp~\citep{koleilat2025biomedcoop} perform under label-noise conditions in both few-shot evaluation and base-to-novel generalization settings.

\para{Setup}
In base$\rightarrow$novel generalization, we corrupt the \emph{16-shot} support set by randomly flipping the labels of $k$ samples per class to other \emph{base} classes. In few-shot evaluation, since training uses all classes, a flipped label can be reassigned to any other class. We vary $k\!\in\!\{2,8,12,16\}$, where ``Num flips = $k$'' means that $k$ of the 16 shots per class are relabeled at random. This protocol reflects realistic medical curation issues (clinician disagreement, ambiguous findings, data-entry errors), and tests whether our geometry-aware component can mitigate noise.

\para{Result of base$\rightarrow$novel generalization}
Table~\ref{tab:flip_robustness_2x2} reports Base/Novel/HM (\%) averaged across 10 datasets. Across all noise levels, our geometry-aware method is consistently more robust to label flips: for \textbf{2} flips, we outperform BiomedCoOp~\citep{koleilat2025biomedcoop} by $(+1.65/+2.10/+1.89)\%$ (Base/Novel/HM); for \textbf{8} flips by $(+2.70/-0.14/+1.27)\%$ (a slight drop on Novel); for \textbf{12} flips by $(+1.61/+0.63/+1.16)\%$; and for the extreme \textbf{16} flips by $(+2.18/+0.60/+1.53)\%$.

\para{Result of few-shot evaluation} Table~\ref{tab:fewshot_labelnoise_singlewide} reports robustness comparisons under different label flip settings. We observe consistently improved robustness across flip settings. Under heavier noise (12 and especially 16 flips), our method shows markedly stronger tolerance than BiomedCoOp~\citep{koleilat2025biomedcoop}.

\para{Why geometry helps}
When some supports are mislabeled, our geometry-aware prior \textit{provides stable, structured targets}, reduces spurious confidence on distant classes, and retains stronger robustness compared to BiomedCoOp~\citep{koleilat2025biomedcoop} as noise increases. Our geometry-aware component does not rely on any training samples, as our class graph $\mathbf{W}$ is \textit{text-driven}. Even with all 16 shots per class flipped (16/16), the prior still provides useful guidance, translating into consistent HM gains over BiomedCoOp~\citep{koleilat2025biomedcoop}. This behavior is desirable in medical settings, where imperfections in annotation are common, but safety performance is required.

\section{Additional hyperparameters and pseudocode}
\label{app:hyperparameters_pseudocode}
Table~\ref{tab:hp_ours} includes the value of hyperparameters used to train the proposed method. Following BiomedCoOp~\citep{koleilat2025biomedcoop}, we select these values using validation set from official datasets.

\paragraph{Pseudocode of proposed method.} We use algorithm~\ref{alg:build-W} to construct the class graph $\mathbf{W}$ and algorithm~\ref{alg:train} for the overall training process. The class graph $\mathbf{W}$ is computed once and kept frozen during training.

\begin{table}[t]
\centering
\begingroup
\scriptsize
\setlength{\tabcolsep}{4pt}
\renewcommand{\arraystretch}{0.98}
\caption{Hyperparameters per dataset and benchmark for our method. \(\lambda_{\mathrm{GAD}}\) and \(\lambda_{\mathrm{LGD}}\) are loss weights; \(\gamma\) is the geometry strength; \(K/P\) is the top-token ratio.}
\label{tab:hp_ours}
\resizebox{0.6\linewidth}{!}{%
\begin{tabular}{l l c c c c}
\toprule
Dataset & Benchmark & $\lambda_{\mathrm{GAD}}$ & $\lambda_{\mathrm{LGD}}$ & $\gamma$ & $K/P$ \\
\midrule
\multirow{2}{*}{BTMRI}    & Base-to-Novel  & 0.50  & 0.01  & 0.80 & 0.50 \\
                          & Few-shot       & 1.75  & 0.05  & 0.50 & 0.40 \\
\midrule
\multirow{2}{*}{BUSI}     & Base-to-Novel  &
\multicolumn{1}{c}{\textemdash} &
\multicolumn{1}{c}{\textemdash} &
\multicolumn{1}{c}{\textemdash} &
\multicolumn{1}{c}{\textemdash} \\
                          & Few-shot       & 0.10  & 0.75  & 0.05 & 0.80 \\
\midrule
\multirow{2}{*}{COVID-QU\mbox{-}Ex} & Base-to-Novel  & 0.50  & 2.25  & 0.05 & 0.50 \\
                                   & Few-shot       & 3.50  & 0.75  & 0.10 & 0.20 \\
\midrule
\multirow{2}{*}{CTKIDNEY} & Base-to-Novel  & 2.00  & 0.25  & 0.90 & 0.30 \\
                          & Few-shot       & 1.25  & 0.10  & 0.70 & 0.03 \\
\midrule
\multirow{2}{*}{DermaMNIST} & Base-to-Novel & 3.50  & 0.10  & 0.10 & 0.30 \\
                            & Few-shot      & 24.00 & 4.00  & 0.60 & 0.10 \\
\midrule
\multirow{2}{*}{Kvasir}   & Base-to-Novel  & 3.50  & 22.00 & 0.30 & 0.03 \\
                          & Few-shot       & 0.50  & 1.00  & 0.01 & 0.01 \\
\midrule
\multirow{2}{*}{CHMNIST}  & Base-to-Novel  & 3.75  & 0.01  & 0.30 & 0.05 \\
                          & Few-shot       & 2.25  & 0.05  & 0.70 & 0.80 \\
\midrule
\multirow{2}{*}{LC25000}  & Base-to-Novel  & 0.25  & 2.00  & 0.40 & 0.60 \\
                          & Few-shot       & 1.00  & 0.05  & 0.50 & 0.05 \\
\midrule
\multirow{2}{*}{RETINA}   & Base-to-Novel  & 0.50  & 0.01  & 0.30 & 0.90 \\
                          & Few-shot       & 0.50  & 1.25  & 0.70 & 0.10 \\
\midrule
\multirow{2}{*}{KneeXray} & Base-to-Novel  & 4.75  & 0.01  & 0.30 & 0.60 \\
                          & Few-shot       & 20.00 & 1.75  & 0.40 & 0.80 \\
\midrule
\multirow{2}{*}{OCTMNIST} & Base-to-Novel  & 0.25  & 0.03  & 0.70 & 0.01 \\
                          & Few-shot       & 4.00  & 3.25  & 0.70 & 0.03 \\
\bottomrule
\end{tabular}%
}
\endgroup
\end{table}

\begin{table}[tb]
\centering
\caption{\textbf{Robustness comparison under label-noise (random label flips) in base$\rightarrow$novel generalization.} Base/Novel/HM accuracies (\%) across 10 datasets with 16 shots over 3 runs. Arrows in the \textbf{AVERAGE} block denote (Ours $-$ BiomedCoOp~\citep{koleilat2025biomedcoop}): green $\uparrow$ = improvement; red $\downarrow$ = drop.}
\label{tab:flip_robustness_2x2}
\begingroup
\scriptsize
\setlength{\tabcolsep}{3.2pt}
\renewcommand{\arraystretch}{1.08}
\begin{minipage}[t]{0.49\linewidth}\centering
\textbf{Num flips = 2}\vspace{2pt}
\resizebox{\linewidth}{!}{%
\begin{tabular}{lrrrrrr}
\toprule
& \multicolumn{3}{c}{\textbf{Ours}} & \multicolumn{3}{c}{\textbf{BiomedCoOp}} \\
\cmidrule(lr){2-4}\cmidrule(lr){5-7}
Dataset & Base & Novel & HM & Base & Novel & HM \\
\midrule
BTMRI       & 81.79 & 96.40 & 88.50 & 79.29 & 95.35 & 86.58 \\
CHMNIST     & 88.34 & 45.70 & 60.24 & 85.88 & 44.64 & 58.98 \\
COVID-QU-Ex & 75.72 & 91.25 & 82.76 & 74.77 & 90.57 & 81.92 \\
CTKIDNEY    & 80.58 & 81.13 & 80.85 & 85.77 & 74.24 & 79.59 \\
DermaMNIST  & 64.08 & 45.03 & 52.89 & 63.80 & 55.74 & 54.75 \\
KneeXray    & 42.16 & 77.01 & 54.49 & 43.54 & 63.14 & 51.54 \\
Kvasir      & 86.39 & 61.28 & 71.70 & 84.94 & 58.00 & 68.93 \\
LC25000     & 89.80 & 95.00 & 92.33 & 93.92 & 93.29 & 91.35 \\
OCTMNIST    & 81.80 & 52.67 & 64.08 & 77.00 & 50.00 & 60.63 \\
RETINA      & 69.25 & 66.25 & 67.72 & 63.46 & 65.14 & 66.76 \\
\midrule
\textbf{AVERAGE} & 75.99 & 71.17 & 73.50 & 74.34 & 69.07 & 71.61 \\
& \multicolumn{1}{c}{\avgup{1.65}} & \multicolumn{1}{c}{\avgup{2.10}} & \multicolumn{1}{c}{\avgup{1.89}} & \multicolumn{3}{c}{} \\
\bottomrule
\end{tabular}%
}
\end{minipage}\hfill
\begin{minipage}[t]{0.49\linewidth}\centering
\textbf{Num flips = 8}\vspace{2pt}
\resizebox{\linewidth}{!}{%
\begin{tabular}{lrrrrrr}
\toprule
& \multicolumn{3}{c}{\textbf{Ours}} & \multicolumn{3}{c}{\textbf{BiomedCoOp}} \\
\cmidrule(lr){2-4}\cmidrule(lr){5-7}
Dataset & Base & Novel & HM & Base & Novel & HM \\
\midrule
BTMRI       & 72.67 & 96.33 & 82.84 & 69.00 & 84.88 & 76.12 \\
CHMNIST     & 86.08 & 44.42 & 58.60 & 85.81 & 39.67 & 54.26 \\
COVID-QU-Ex & 63.52 & 91.56 & 75.01 & 65.06 & 91.37 & 76.00 \\
CTKIDNEY    & 71.55 & 81.31 & 76.12 & 73.76 & 77.87 & 75.76 \\
DermaMNIST  & 62.46 & 47.19 & 53.76 & 50.89 & 61.67 & 55.76 \\
KneeXray    & 41.25 & 64.72 & 50.39 & 41.44 & 66.79 & 51.15 \\
Kvasir      & 85.33 & 58.78 & 69.61 & 82.78 & 55.17 & 66.21 \\
LC25000     & 86.97 & 95.33 & 90.96 & 85.44 & 95.29 & 90.10 \\
OCTMNIST    & 74.93 & 49.60 & 59.69 & 70.00 & 50.07 & 58.59 \\
RETINA      & 65.03 & 61.57 & 63.25 & 58.03 & 69.45 & 63.23 \\
\midrule
\textbf{AVERAGE} & 70.98 & 69.08 & 70.02 & 68.28 & 69.22 & 68.75 \\
& \multicolumn{1}{c}{\avgup{2.70}} & \multicolumn{1}{c}{\avgdown{0.14}} & \multicolumn{1}{c}{\avgup{1.27}} & \multicolumn{3}{c}{} \\
\bottomrule
\end{tabular}%
}
\end{minipage}
\vspace{8pt}
\begin{minipage}[t]{0.49\linewidth}\centering
\textbf{Num flips = 12}\vspace{2pt}
\resizebox{\linewidth}{!}{%
\begin{tabular}{lrrrrrr}
\toprule
& \multicolumn{3}{c}{\textbf{Ours}} & \multicolumn{3}{c}{\textbf{BiomedCoOp}} \\
\cmidrule(lr){2-4}\cmidrule(lr){5-7}
Dataset & Base & Novel & HM & Base & Novel & HM \\
\midrule
BTMRI       & 64.00 & 95.45 & 76.65 & 59.20 & 84.04 & 69.45 \\
CHMNIST     & 81.30 & 44.99 & 57.91 & 82.60 & 37.41 & 51.50 \\
COVID-QU-Ex & 49.50 & 91.19 & 64.18 & 58.10 & 90.76 & 70.85 \\
CTKIDNEY    & 53.90 & 78.99 & 64.08 & 51.20 & 75.99 & 61.19 \\
DermaMNIST  & 62.40 & 53.71 & 57.72 & 52.50 & 65.31 & 58.21 \\
KneeXray    & 42.90 & 65.45 & 51.80 & 42.70 & 75.43 & 54.56 \\
Kvasir      & 84.80 & 55.56 & 67.14 & 82.00 & 49.84 & 61.98 \\
LC25000     & 79.50 & 89.02 & 83.99 & 76.90 & 94.26 & 84.71 \\
OCTMNIST    & 68.70 & 50.00 & 57.87 & 71.00 & 48.20 & 57.42 \\
RETINA      & 59.20 & 70.08 & 64.21 & 53.80 & 66.93 & 59.66 \\
\midrule
\textbf{AVERAGE} & 64.60 & 69.44 & 66.94 & 63.00 & 68.82 & 65.78 \\
& \multicolumn{1}{c}{\avgup{1.61}} & \multicolumn{1}{c}{\avgup{0.63}} & \multicolumn{1}{c}{\avgup{1.16}} & \multicolumn{3}{c}{} \\
\bottomrule
\end{tabular}%
}
\end{minipage}\hfill
\begin{minipage}[t]{0.49\linewidth}\centering
\textbf{Num flips = 16}\vspace{2pt}
\resizebox{\linewidth}{!}{%
\begin{tabular}{lrrrrrr}
\toprule
& \multicolumn{3}{c}{\textbf{Ours}} & \multicolumn{3}{c}{\textbf{BiomedCoOp}} \\
\cmidrule(lr){2-4}\cmidrule(lr){5-7}
Dataset & Base & Novel & HM & Base & Novel & HM \\
\midrule
BTMRI       & 51.42 & 96.73 & 67.15 & 45.63 & 89.82 & 60.52 \\
CHMNIST     & 78.10 & 41.89 & 54.53 & 79.12 & 39.89 & 53.04 \\
COVID-QU-Ex & 43.56 & 91.34 & 58.99 & 49.31 & 89.77 & 63.65 \\
CTKIDNEY    & 55.91 & 72.79 & 63.24 & 50.17 & 74.04 & 59.81 \\
DermaMNIST  & 62.22 & 62.71 & 62.46 & 49.11 & 82.44 & 61.55 \\
KneeXray    & 41.08 & 71.53 & 52.19 & 41.15 & 49.52 & 44.95 \\
Kvasir      & 82.28 & 55.28 & 66.13 & 80.17 & 54.11 & 64.61 \\
LC25000     & 80.24 & 93.15 & 86.21 & 75.24 & 92.24 & 82.88 \\
OCTMNIST    & 62.67 & 49.20 & 55.12 & 64.07 & 51.67 & 57.21 \\
RETINA      & 48.71 & 67.45 & 56.57 & 50.40 & 72.60 & 59.50 \\
\midrule
\textbf{AVERAGE} & 60.62 & 70.21 & 65.06 & 58.44 & 69.61 & 63.54 \\
& \multicolumn{1}{c}{\avgup{2.18}} & \multicolumn{1}{c}{\avgup{0.60}} & \multicolumn{1}{c}{\avgup{1.53}} & \multicolumn{3}{c}{} \\
\bottomrule
\end{tabular}%
}
\end{minipage}

\endgroup
\end{table}

\begin{table}[tb]
\centering
\caption{\textbf{Robustness comparison under label noise (random label flips) in few-shot evaluation.}
Average accuracy (\%) across 11 datasets under different numbers of random label flips (16-shot), averaged over 3 runs.}
\label{tab:fewshot_labelnoise_singlewide}
\begingroup
\scriptsize
\setlength{\tabcolsep}{5pt}
\renewcommand{\arraystretch}{1.05}
\resizebox{\textwidth}{!}{%
\begin{tabular}{l r c r c r c r c}
\toprule
& \multicolumn{2}{c}{\textbf{Num flips = 2}}
& \multicolumn{2}{c}{\textbf{Num flips = 8}}
& \multicolumn{2}{c}{\textbf{Num flips = 12}}
& \multicolumn{2}{c}{\textbf{Num flips = 16}} \\
\cmidrule(lr){2-3}\cmidrule(lr){4-5}\cmidrule(lr){6-7}\cmidrule(lr){8-9}
\textbf{Dataset}
& \textbf{Ours} & \textbf{BiomedCoOp}
& \textbf{Ours} & \textbf{BiomedCoOp}
& \textbf{Ours} & \textbf{BiomedCoOp}
& \textbf{Ours} & \textbf{BiomedCoOp} \\
\midrule
BTMRI       & 82.43 & 81.11 & 79.15 & 78.22 & 74.41 & 76.86 & 73.11 & 73.07 \\
BUSI        & 67.09 & 65.68 & 63.98 & 63.14 & 65.54 & 62.29 & 62.01 & 57.06 \\
CHMNIST     & 78.64 & 78.50 & 77.26 & 77.52 & 74.64 & 73.85 & 75.22 & 75.31 \\
COVID-QU-Ex & 77.95 & 77.76 & 77.56 & 77.42 & 77.06 & 76.86 & 74.83 & 75.85 \\
CTKIDNEY    & 80.78 & 82.81 & 76.61 & 79.31 & 78.47 & 75.51 & 72.89 & 69.53 \\
DermaMNIST  & 65.65 & 62.76 & 66.22 & 62.14 & 64.42 & 61.61 & 64.29 & 61.39 \\
KneeXray    & 41.49 & 39.72 & 40.82 & 39.41 & 40.52 & 39.57 & 40.60 & 39.90 \\
Kvasir      & 80.14 & 78.67 & 77.94 & 77.75 & 77.44 & 76.06 & 76.00 & 74.81 \\
LC25000     & 90.71 & 91.85 & 87.83 & 89.16 & 85.00 & 83.99 & 83.97 & 86.95 \\
OCTMNIST    & 67.90 & 66.90 & 66.60 & 64.83 & 66.10 & 61.87 & 65.17 & 59.57 \\
RETINA      & 60.81 & 61.25 & 54.13 & 55.65 & 49.97 & 53.58 & 47.08 & 49.34 \\
\midrule
\textbf{AVERAGE}
& \textbf{72.14} & 71.55
& \textbf{69.83} & 69.50
& \textbf{68.51} & 67.46
& \textbf{66.83} & 65.71 \\
\\[-6pt]
& \multicolumn{1}{c}{\avgup{0.60}} & \multicolumn{1}{c}{}
& \multicolumn{1}{c}{\avgup{0.32}} & \multicolumn{1}{c}{}
& \multicolumn{1}{c}{\avgup{1.05}} & \multicolumn{1}{c}{}
& \multicolumn{1}{c}{\avgup{1.13}} & \multicolumn{1}{c}{} \\
\bottomrule
\end{tabular}%
}
\endgroup
\end{table}

\begin{algorithm}[h]
\caption{Build class graph $\mathbf{W}$}
\label{alg:build-W}
\small
\begin{algorithmic}[0]
\STATE \makebox[3.8em][l]{\textbf{Input:}}%
\parbox[t]{\dimexpr\linewidth-3.8em\relax}{%
Frozen text encoder $g^{\text{text}}$; class set $\mathcal{C}=\{1,\dots,C\}$.\\ Note that in few-shot evaluation, $\mathcal{C}$ means using all classes; whereas in base-to-novel generalization, $\mathcal{C}$ means using only base classes.
}
\STATE \hspace*{3.8em}\parbox[t]{\dimexpr\linewidth-3.8em\relax}{%
Prompt bank $\{\mathcal{P}_c\}_{c=1}^C$ (e.g., 50 prompts per class); graph sharpness $\alpha>0$.
}
\STATE \makebox[3.8em][l]{\textbf{Output:}}%
\parbox[t]{\dimexpr\linewidth-3.8em\relax}{%
class graph $\mathbf{W}\in\mathbb{R}^{C\times C}$ (kept frozen).
}
\FOR{$c \in \{1,\dots,C\}$}
  \STATE \textbf{Text prototype for class $c$} (average of frozen prompt features)
  \STATE $\mathbf{t}_c \leftarrow \frac{1}{|\mathcal{P}_c|}\sum\limits_{p\in \mathcal{P}_c}
     \mathrm{normalize}\!\big(g^{\text{text}}(p)\big)$ \algCmt{L2-normalize}
\ENDFOR
\STATE $\mathbf{T} \leftarrow [\,\mathbf{t}_1,\dots,\mathbf{t}_C\,]^\top \in \mathbb{R}^{C\times d}$
\STATE $\mathbf{S} \leftarrow \mathbf{T}\mathbf{T}^\top \in [-1,1]^{C\times C}$
\FOR{$c \in \{1,\dots,C\}$}
  \STATE \textit{Row-wise softmax to get non-negative weights that sum to $1$}
\STATE $\mathbf{W}_{c:} \leftarrow \softmax\!\big(\alpha\,\mathbf{S}_{c:}\big)$ \algCmt{$\alpha\uparrow$: sharper neighborhoods;\ $\alpha\downarrow$: more spread}
\ENDFOR
\STATE $\mathbf{W} \leftarrow \texttt{torch.stack}(\,[\mathbf{W}_{c:}]_{c=1}^{C},\, \texttt{dim=0})$
\STATE \textbf{return} $\mathbf{W}$
\end{algorithmic}
\end{algorithm}

\begin{algorithm}[t]
\setlength{\abovecaptionskip}{2pt}
\setlength{\belowcaptionskip}{2pt}
\par\noindent\hrule height 0.9pt
\vspace{1pt}
\caption{Training loop of proposed method in PyTorch-like style}
\label{alg:train}
\vspace{1pt}
\par\noindent\hrule height 0.4pt
\vspace{2pt}
\footnotesize
\begin{algorithmic}[0]
\STATE \textbf{Input:} $\mathbf{T}\in\mathbb{R}^{C\times d}$, $\mathbf{W}$ (Alg.~\ref{alg:build-W}), $\tilde{\mathbf{T}}\in\mathbb{R}^{C\times d}$ (frozen);
$\tau,T,\gamma,K,\lambda_1,\lambda_2,\lambda_3$; context $\mathbf{C}$; dataset $\mathcal{D}$; optimizer on $\mathbf{C}$ only.
\STATE \textbf{Output:} Updated context $\mathbf{C}$ (encoders and $\mathbf{W}$ remain frozen).
\FOR{each epoch}
  \FOR{mini-batch $\mathcal{B}=\{(\mathbf{x},y)\}$}
    \STATE $\hat{\mathbf{T}} \leftarrow g^{\text{text}}(\text{class prompts with context }\mathbf{C})$ \algCmtR{student text (L2-normalized rows)}
    \STATE $\mathbf{v}(\mathbf{x}) \leftarrow g^{\text{img}}(\mathbf{x}).\texttt{global\_token}()$ \algCmtR{global \IMG{} token, L2-normalized}

    \STATE $\mathbf{z}_{\text{stu}}(\mathbf{x}) \leftarrow \tau \,\hat{\mathbf{T}}\,\mathbf{v}(\mathbf{x})$
    \STATE $\tilde{\mathbf{z}}(\mathbf{x}) \leftarrow \tau \,\tilde{\mathbf{T}}\,\mathbf{v}(\mathbf{x})$

    \STATE $\boldsymbol{\ell}_{\text{stu}}(\mathbf{x}) \leftarrow \log\!\softmax\!\big(\mathbf{z}_{\text{stu}}(\mathbf{x}) / T\big)$
    \STATE $\boldsymbol{\ell}_{\text{tea}}(\mathbf{x}) \leftarrow \log\!\softmax\!\big(\tilde{\mathbf{z}}(\mathbf{x}) / T\big)$

    \STATE $\boldsymbol{\ell}_{\text{tea}}^{\star}(\mathbf{x})
      \leftarrow (1-\gamma)\,\boldsymbol{\ell}_{\text{tea}}(\mathbf{x})
        + \gamma\,\boldsymbol{\ell}_{\text{tea}}(\mathbf{x})\,\mathbf{W}$ \algCmtR{geometry-aware teacher (global)}

    \STATE \textbf{Global geometry-aware distillation (GAD)}
    \STATE $\mathcal{L}_{\GAD} \leftarrow \frac{1}{|\mathcal{B}|}\sum\limits_{\mathbf{x}\in\mathcal{B}}
        \mathrm{KL}\!\big(\boldsymbol{\ell}_{\text{tea}}^{\star}(\mathbf{x})\,\|\,\boldsymbol{\ell}_{\text{stu}}(\mathbf{x})\big)$

    \STATE $\mathbf{Z}(\mathbf{x}) \leftarrow g^{\text{img}}(\mathbf{x}).\texttt{patch\_tokens}()$ \algCmtR{patch tokens, shape $[P,d]$}
    \STATE $\mathbf{S}_{\text{stu}}(\mathbf{x}) \leftarrow \tau\,\mathbf{Z}(\mathbf{x})\,\hat{\mathbf{T}}^\top$ \algCmtR{patch logits}
    \STATE $\mathbf{S}_{\text{tea}}(\mathbf{x}) \leftarrow \tau\,\mathbf{Z}(\mathbf{x})\,\tilde{\mathbf{T}}^\top$

    \STATE $\mathbf{L}_{\text{stu}}(\mathbf{x}) \leftarrow \log\!\softmax\!\big(\mathbf{S}_{\text{stu}}(\mathbf{x})/T\big)$
    \STATE $\mathbf{L}_{\text{tea}}(\mathbf{x}) \leftarrow \log\!\softmax\!\big(\mathbf{S}_{\text{tea}}(\mathbf{x})/T\big)$

    \STATE $\mathbf{L}_{\text{tea}}^{\star}(\mathbf{x})
      \leftarrow (1-\gamma)\,\mathbf{L}_{\text{tea}}(\mathbf{x})
        + \gamma\,\mathbf{L}_{\text{tea}}(\mathbf{x})\,\mathbf{W}$ \algCmtR{geometry-aware teacher (patch)}

    \STATE \textit{(Shown as loop for clarity; implement in batch mode.)}
    \STATE Initialize accumulator $A \leftarrow 0$
    \FOR{each $(\mathbf{x},y)\in\mathcal{B}$}
      \STATE $s_n \leftarrow \cos\!\big(\mathbf{z}_n(\mathbf{x}), \tilde{\mathbf{t}}_y\big)$ for $n=1,\dots,P$ \algCmtR{align w/ teacher GT text}
      \STATE $\mathcal{I}_K(\mathbf{x},y) \leftarrow \TopK\big(\{s_n\}_{n=1}^P, K\big)$
      \STATE $u^{\text{stu}} \leftarrow \big(\mathbf{L}_{\text{stu}}(\mathbf{x})\,\mathbf{e}_y\big)\big|_{\mathcal{I}_K(\mathbf{x},y)}$
      \STATE $u^{\text{tea}\star} \leftarrow \big(\mathbf{L}_{\text{tea}}^{\star}(\mathbf{x})\,\mathbf{e}_y\big)\big|_{\mathcal{I}_K(\mathbf{x},y)}$
      \STATE $A \leftarrow A + \mathrm{KL}\!\big(u^{\text{tea}\star} \,\|\, u^{\text{stu}}\big)$
    \ENDFOR

    \STATE \textbf{Label-guided Geometry Distillation (LGD)}
    \STATE $\mathcal{L}_{\LGD} \leftarrow \frac{1}{|\mathcal{B}|}\,A$

    \STATE $\mathbf{p}_{\text{stu}}(\mathbf{x}) \leftarrow \softmax\!\big(\mathbf{z}_{\text{stu}}(\mathbf{x})\big)$
    \STATE $\mathcal{L}_{\text{CE}} \leftarrow \frac{1}{|\mathcal{B}|}\sum\limits_{(\mathbf{x},y)\in\mathcal{B}} \big[-\log \mathbf{p}_{\text{stu}}(\mathbf{x})_y\big]$
    \STATE $\mathcal{L}_{\text{SCCM}} \leftarrow \frac{1}{C}\,\|\hat{\mathbf{T}}-\mathbf{T}\|_F^2$

    \STATE \textbf{Overall training loss}
    \STATE $\mathcal{L} \leftarrow \mathcal{L}_{\text{CE}} + \lambda_1 \mathcal{L}_{\text{SCCM}} + \lambda_2 \mathcal{L}_{\GAD} + \lambda_3 \mathcal{L}_{\LGD}$
    \STATE \texttt{optimizer.zero\_grad()}
    \STATE backprop $\nabla_{\mathbf{C}}\mathcal{L}$; \texttt{optimizer.step()} \algCmtR{encoders \& $\mathbf{W}$ frozen}
  \ENDFOR
\ENDFOR

\end{algorithmic}
\vspace{4pt}\hrule\vspace{2pt}
\end{algorithm}

\end{document}